\documentclass[11pt]{article}

\usepackage[margin=1.13in]{geometry}

\usepackage[square]{natbib}
\usepackage{graphicx}
\usepackage[caption=false]{subfig}
\usepackage{amsmath, amssymb}

\newcommand{\MIC}{\mbox{MIC}}

\newenvironment{definition}[1][Definition]{\begin{trivlist}
\item[\hskip \labelsep {\bfseries #1}]}{\end{trivlist}}

\begin{document}

\title{Equitability Analysis of the Maximal Information Coefficient, with Comparisons}

\author{\textbf{David N.\ Reshef}\footnote{These authors contributed equally to this work.} \\ Department of Electrical Engineering and Computer Science\\ Harvard-MIT Division of Heath Sciences and Technology \\ Massachusetts Institute of Technology \\ dnreshef@mit.edu \and
\textbf{Yakir A. Reshef}\footnotemark[\value{footnote}] \\ Harvard-MIT Division of Heath Sciences and Technology \\ Massachusetts Institute of Technology \\ yakirr@mit.edu \and
\textbf{Michael M. Mitzenmacher}\footnote{These authors contributed equally to this work} \\ School of Engineering and Applied Sciences \\ Harvard University \\ michaelm@eecs.harvard.edu \and
\textbf{Pardis C. Sabeti}\footnotemark[\value{footnote}] \\ Department of Organismic and Evolutionary Biology\\ Broad Institute of MIT and Harvard \\ Harvard University \\ psabeti@oeb.harvard.edu}

\maketitle

\begin{abstract}
A measure of dependence is said to be \emph{equitable} if it gives similar scores to equally noisy relationships of different types. Equitability is important in data exploration when the goal is to identify a relatively small set of strongest associations within a dataset as opposed to finding as many non-zero associations as possible, which often are too many to sift through. Thus an equitable statistic, such as the maximal information coefficient (MIC), can be useful for analyzing high-dimensional data sets. Here, we explore both equitability and the properties of MIC, and discuss several aspects of the theory and practice of MIC.  We begin by presenting an intuition behind the equitability of MIC through the exploration of the maximization and normalization steps in its definition. We then examine the speed and optimality of the approximation algorithm used to compute MIC, and suggest some directions for improving both. Finally, we demonstrate in a range of noise models and sample sizes that MIC is more equitable than natural alternatives, such as mutual information estimation and distance correlation.
\end{abstract}

\section{Introduction}

In \citet{MINE}, the authors introduce \emph{equitability}: a measure of dependence is said to be equitable if it gives similar scores to relationships with similar noise levels.  Equitability is important in exploration of high-dimensional data sets, where there can be upwards of thousands of pairwise relationships to consider, and there is no a priori reason to prefer finding certain types of relationships over others.  By sorting relationships according to an equitable measure, one hopes to find important patterns of any type for further examination.

Without equitability, entire classes of relationships could be missed, as scores for those relationships might be dominated by those of other classes of relationships.  It is important to emphasize that in the setting we focus on here \--- data exploration \--- we are not focused on determining with maximal power the existence or non-existence of relationships. Rather, the overwhelming number of dimensions, and therefore potential relationships, in our data set forces us to prioritize which of the possibly many significant relationships should be examined first. Given the increasing dimensionality of available data sets, feature selection and dimensionality reduction tasks such as this are becoming increasingly important \citep{guyon2003introduction, hastie2009elements, roweis2000nonlinear, tenenbaum2000global}.

Reshef et al. also introduce a new measure of dependence, the Maximal Information Coefficient (MIC), and show using simulated data that MIC is highly equitable. Comparisons with other current methods \--- including mutual information estimation, distance correlation, the Spearman correlation coefficient, principal curve-based methods, and maximal correlation \--- demonstrate that they comparatively behave significantly less equitably.  MIC is therefore useful for identifying a subset of strongest associations in a data set that contains too many significant associations to sift through manually. MIC has been employed in fields such as genomics \citep{das2012genome,riccadonna2012dtw}, proteomics \citep{pang2012multi}, the microbiome \citep{koren2012host}, sensing \citep{sagl2012ubiquitous}, vaccine design \citep{anderson2012ranking}, and clinical data analysis \citep{wang2012clinical, lin2012maximal}.

Although the concept of equitability and MIC were introduced in \citet{MINE}, more work remains to explore both equitability and the properties of MIC. Here, we examine in detail some important aspects of the theory and practice of MIC: the utility of maximization and normalization in the definition of MIC, the effects of the parameters used in the approximation algorithm for computing MIC on the run-time and accuracy of the algorithm, the effect on equitability of using an approximation algorithm to compute MIC, and the tradeoff between equitability and power through a comparison with the distance correlation of Szekely and Rizzo \citep{szekely2009brownian}.

Given that MIC is based on mutual information but is not itself a mutual information estimator, a natural question is whether MIC is itself truly necessary, or whether mutual information estimation could be more directly applied to provide an equitable measure of dependence.  In this work we address this question as well, expanding significantly on the results of \citet{MINE} by performing an in-depth comparison to mutual information estimation using a range of smoothing parameters and a large set of test functions, noise models, and sample sizes. We find that, while there are a few regimes under which mutual information estimation performs comparably to MIC, MIC is more equitable than mutual information under almost all the noise models we considered, as well as under any noise model we considered at sample sizes below $n=5000$.

\section{Preliminaries}
\label{sec:Preliminaries}

Roughly, a measure of dependence is equitable if relationships that are similarly noisy receive similar scores, regardless of relationship type. As noted in \citet{MINE}, equitability is hard to define rigorously for non-functional relationships. However, consider a setting where the data take the form $(X + N_x, f(X) + N_y)$ where $X$, $N_x$, and $N_y$ are distributed according to some predetermined model (e.g. $X$ is uniformly distributed on $[0,1]$ and $N_x$ and $N_y$ are uniformly distributed on a small interval and independent of each other and of $X$). This setting corresponds to sampling in which both coordinates are subject to noisy measurements.  Here equitability has a clear interpretation: a measure of dependence $\delta$ is equitable to the extent that the $R^2$ of $(X + N_x, f(X) + N_y)$ with respect to the function $f$ depends only on the score assigned by $\delta$ to $(X+N_x, f(X) + N_y)$ (not on $f$), and vice versa. This setting shall be our focus in this paper.

We recall the definition of the Maximal Information Coefficient (MIC) from \citet{MINE}.
\begin{definition}
\label{def:MIC}
Let $D$ be a set of ordered pairs. For a grid $G$, let $D|_G$ denote the probability distribution induced by the data $D$ on the cells of $G$, and let $I(-)$ denote mutual information. Let $I^*(D,x,y) = \max_G I(D|_G)$, where the maximum is taken over all $x$-by-$y$ grids $G$ (possibly with empty rows/columns). MIC is defined as
$$\MIC(D) = \max_{xy < B(|D|)}\frac{I^*(D,x,y)}{\log_2 \min\{x,y\}}$$
Where $B$ is a growing function satisfying $B(n) = o(n)$.
\end{definition}

\begin{figure}
	\centering
	\subfloat {\includegraphics[trim = 0in 2in 0in 0in, clip, height = 0.7\textheight]{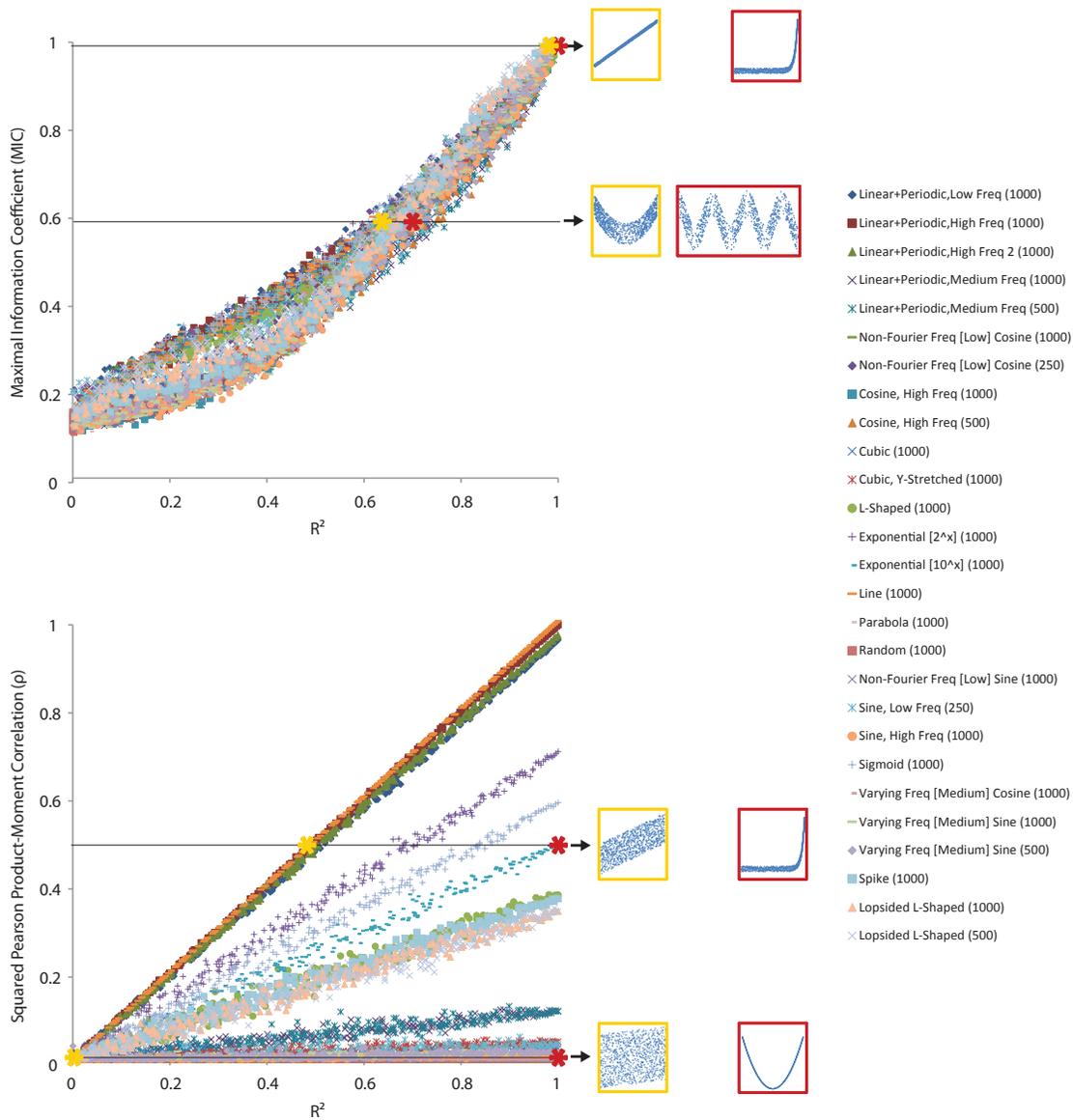}}
	\caption[Equitability of MIC and the Pearson product-moment correlation ($\rho$) on noisy functional relationships]{The equitability of MIC and the Pearson product-moment correlation ($\rho$) on a test suite of noisy functional relationships. The top plot contains the MIC scores of 27 different functional relationships with varying sample sizes and increasing amounts of noise plotted against the coefficient of determination ($R^2$) of each relationship relative to the noiseless function used to generate it. Noise was generated using noise model 1 (points spaced evenly along the curve described by the function, uniform vertical noise). Thumbnails shown to the right of the plot show relationships that received identical MIC scores.  The plot shows that MIC is relatively equitable: it gives similar scores to similarly noisy relationships, independent of relationship type.  As expected, the (squared) Pearson product-moment correlation, which is designed to detect linear dependence, is not equitable, giving widely varying scores to different types of relationships with similar noise levels. Function names correspond to those in Table~\ref{table:FctTestSuite}.}
\label{EquitabilityIntro}
\end{figure}

In \citet{MINE}, the authors heuristically suggest $B(n) = n^{0.6}$ as a default setting, though we will show here that for larger sample size regimes different settings of $B(n)$ speed up the computation of MIC without sacrificing much in the way of equitability. An approximation of MIC can be computed efficiently using dynamic programming \citep{MINE};  when we refer to computing MIC, we mean according to this approximation method unless otherwise stated, and we use $I^*$ to denote the computed values instead of the idealized values henceforth.  Figure~\ref{EquitabilityIntro}, based on Figure 2b from \citet{MINE}, demonstrates the equitability of MIC on a suite of functional relationships.

In evaluating the equitability of MIC or other measures of dependence, it is important to consider a range of sampling and noise models, as these could affect the equitability of various measures of dependence differently.  In previous work, simulations showed that MIC is more equitable than existing methods across a range of function types, for four basic noise models, and for sample sizes ranging from $n=250$ to $n=1000$ \citep{MINE}. To further characterize the equitability of MIC, in this work we extend the comparisons to larger sample sizes and consider additional noise models.  However, our goal is not just to contrast MIC with existing schemes, but to offer some insight into why it performs better in many settings than other possibly reasonable alternatives.

For the purpose of our analysis, we utilize six different sampling/noise models that may be found in real data sets. Each noise model is specified by: \\
\\
\begin{itemize}
\item Whether data points are chosen equally spaced along the curve described by the function in question (models 1-3) or equally spaced along the $x$-axis range (models 4-6)
\item Whether noise is added in the $y$ coordinate (models 1,4), in the $x$ coordinate (models 3, 6), or in both (models 2, 5).  The noise added is uniform over an interval (in models 2 and 5 the same interval is used for both noise distributions), with the interval increasing in size over our trials to provide a diversity in the added noise, as we describe in our results.
\end{itemize}
Tables~\ref{table:NoiseModels} and~\ref{table:FctTestSuite}, respectively, summarize the sampling/noise models and contain the definitions of the functions used to assess the equitability of various methods throughout this paper.

\begin{table}
\small \begin{tabular}{|c|c|c|}
	\hline
	\textbf{Noise added in}				& \textbf{Points sampled equally spaced along}	& \textbf{Points sampled equally spaced}	\\
					& \textbf{curve described by function}			& \textbf{along $x$-axis range}		\\ \hline
	$y$-coordinate								& Noise Model 1					& Noise Model 4				\\ \hline
	$x$, $y$-coordinates						& Noise Model 2					& Noise Model 5				\\ \hline
	$x$-coordinate								& Noise Model 3					& Noise Model 6				\\ 
	\hline
\end{tabular}
\caption[Summary of sampling/noise models used.]{Summary of sampling/noise models used.  Sampling/noise models are specified by (1) how points are sampled from the distribution defined by the function, and (2) the coordinates to which noise is added.}
\label{table:NoiseModels}
\end{table}

\begin{table}
\centering
\begin{tabular}{|l|ll|}
\hline
\textbf{Function Name} 			& \textbf{Definition}														& \\
\hline
Linear+Periodic, Low Freq			& $y = \frac{1}{5}\sin(4(2x-1)) + \frac{11}{10}(2x-1)$ 							& $x \in [0,1]$ \\
\hline
Linear+Periodic, Medium Freq		& $y = \sin(10\pi x) + x$													& $x \in [0,1]$ \\
\hline
Linear+Periodic, High Freq			& $y = \frac{1}{10}\sin(10.6(2x-1)) + \frac{11}{10}(2x-1)$						& $x \in [0,1]$ \\
\hline
Linear+Periodic, High Freq 2			& $y = \frac{1}{5}\sin(10.6(2x-1)) + \frac{11}{10}(2x-1)$						& $x \in [0,1]$ \\
\hline
Non-Fourier Freq [Low] Cosine		& $y = \cos(7\pi x)$														& $x \in [0,1]$ \\
\hline
Cosine, High Freq					& $y = \cos(14\pi x)$														& $x \in [0,1]$ \\
\hline
Cubic							& $y = 4x^3 + x^2 - 4x$													& $x \in [-1.3,1.1]$ \\
\hline
Cubic, Y-stretched					& $y = 41(4x^3 + x^2 - 4x)$												& $x \in [-1.3,1.1]$ \\
\hline
L-shaped							& $y = \begin{cases}
									x/99		& \text{if } x \leq \frac{99}{100} \\
									1		& \text{if } x > \frac{99}{100}
								\end{cases} $																& $x \in [0,1]$ \\
\hline
Exponential [$2^x$]				& $y = 2^x$																& $x \in [0,10]$ \\
\hline
Exponential [$10^x$]				& $y = 10^x$																& $x \in [0,10]$ \\
\hline
Line								& $y = x$																& $x \in [0,1]$ \\
\hline
Parabola							& $y = 4x^2$																& $x \in [-\frac{1}{2},\frac{1}{2}]$\\
\hline
Random							& random number generator													& $x \in [0,1]$ \\
\hline
Non-Fourier Freq [Low] Sine			& $y = \sin(9\pi x)$														& $x \in [0,1]$ \\
\hline
Sine, Low Freq						& $y = \sin(8\pi x)$														& $x \in [0,1]$\\
\hline
Sine, High Freq					& $y = \sin(16\pi x)$														& $x \in [0,1]$\\
\hline
Sigmoid							& $y = \begin{cases}
									0								& \text{if } x \leq \frac{49}{100} \\
									50(x-\frac{1}{2}) + \frac{1}{2}		& \text{if } \frac{49}{100} \leq x \leq \frac{51}{100} \\
									1								& \text{if } x > \frac{51}{100}
								\end{cases} $																& $x \in [0,1]$ \\
\hline
Varying Freq [Medium] Cosine		& $y = \sin(5\pi x (1+x)) $													& $x \in [0,1]$ \\
\hline
Varying Freq [Medium] Sine			& $y = \sin(6\pi x (1+x)) $													& $x \in [0,1]$ \\
\hline
Spike							& $y = \begin{cases}
									20 						& \text{if } x < \frac{1}{20} \\
									-18x + \frac{19}{10}		& \text{if } \frac{1}{20} \leq x < \frac{1}{10} \\
									-\frac{x}{9} + \frac{1}{9}	& \text{if } x \geq \frac{1}{10}
								\end{cases} $																& $x \in [0,1]$ \\
\hline
Lopsided L-shaped					& $y = \begin{cases}
									200x 						& \text{if } x < \frac{1}{200} \\
									-198x + \frac{199}{100}		& \text{if } \frac{1}{200} \leq x < \frac{1}{100} \\
									-\frac{x}{99} + \frac{1}{99}	& \text{if } x \geq \frac{1}{100}
								\end{cases} $																& $x \in [0,1]$ \\
\hline
\end{tabular}
\caption[Definitions of the functions used to analyze the equitability of various measures of dependence.]{Definitions of the functions used to analyze the equitability of various measures of dependence.}
\label{table:FctTestSuite}
\end{table}

\section{An Intuition behind the Equitability of MIC}
\label{sec:intuition}

Before examining the properties of MIC in relation to those of other methods, we first explore the features of MIC itself.  We do so by omitting specific features from the definition of MIC and seeing how the resulting statistic behaves.

As seen in Equation~\ref{def:MIC}, MIC contains both a maximization and a normalization step, which together maximize a normalized variant of mutual information over a set of potential grids. We first consider a variation on MIC that omits the maximization step. That is, rather than considering all grids at a given resolution and computing the maximal possible mutual information achieved by any of them, it simply uses the mutual information achieved by an (adaptive) equipartition at each grid resolution.
\begin{definition}
\label{def:MIC_1}
Let $D$ be a set of ordered pairs. Let $E(D,x,y)$ be an $x$-by-$y$ equipartition of $D$; that is, the rows of $E(D,x,y)$ each contain the same number of points\footnote{It is possible that $x$ and $y$ will not divide $|D|$, or $D$ may contain points with identical $x$- or $y$-values. In these cases, we can think of $E(D,x,y)$ as being a grid that is closest to an equipartition rather than an actual equipartition.} of $D$, and the same is true of the columns. Let $I^E(D,x,y) = I(D|_{E(D,x,y)})$. Then $\MIC_1$, the variant of MIC that lacks the step that maximizes over grid partitions, is defined by
$$\MIC_1(D) =  \max_{xy < B(|D|)}\frac{I^E(D,x,y)}{\log_2 \min\{x,y\}}$$
\end{definition}

Second, we consider a variation of MIC that omits the normalization by $\log_2 \min\{x,y\}$.  The choice of $\log_2 \min\{x,y\}$ in the definition of MIC is based on the fact that $\log_2 \min\{x,y\}$ is an upper bound on the maximum possible mutual information of an $x$-by-$y$ grid that is uniform over all such grids.  Hence, this normalization gives values between 0 and 1, and allows for comparisons between grids of different resolutions.\footnote{Instead of normalizing by $\log_2 \min\{x; y\}$, we might also consider normalizing by $\min\{H(X);H(Y)\}$, where $X$ and $Y$ denote the marginal distributions in each dimension of a grid used in $I^*$. This normalization would also have the property that MIC lies between 0 and 1. To see why it is less appealing, consider the family of distributions over the unit square defined as follows: for $0 \leq \alpha \leq 1$, let $D_\alpha$ be a random variable that is uniformly distributed over the region $[0,\alpha] \times [0,\alpha] \cup [\alpha, 1] \times [\alpha, 1]$. The MIC of $D_\alpha$ is $H(\alpha)$ where $H(-)$ is the binary entropy function; however, if the above variant of MIC were used, $D_\alpha$ would have an MIC of 1 for $0 < \alpha < 1$, while $D_0$ and $D_1$, being joint distributions exhibiting statistical independence, would receive scores of 0. In the first case MIC is a continuous function of alpha, while in the latter it is not.}
\begin{definition}
\label{def:MIC_2}
Let $D$ be a set or ordered pairs, and let $I^*$ be as in Definition~\ref{def:MIC}. Then $\MIC_2$, the variant of MIC that omits the normalization step, is defined by
$$\MIC_2(D) = \max_{xy < B(|D|)} I^*(D,x,y)$$
\end{definition}

Finally, we consider a variant of MIC with neither maximization nor normalization, $\MIC_3$.
\begin{definition}
\label{def:MIC_3}
Let $D$ be a set of ordered pairs, and let $I^E$ be as in Definition~\ref{def:MIC_1}. Then $\MIC_3$, the variant of MIC that lacks both the maximization and normalization steps, is defined by:
$$\MIC_3(D) = \max_{xy < B(|D|)} I^E(D,x,y)$$
\end{definition}

\begin{figure}
        \centering
        \includegraphics[trim = 0in 0in 0.125in 0in, clip, width = 0.78\textwidth]{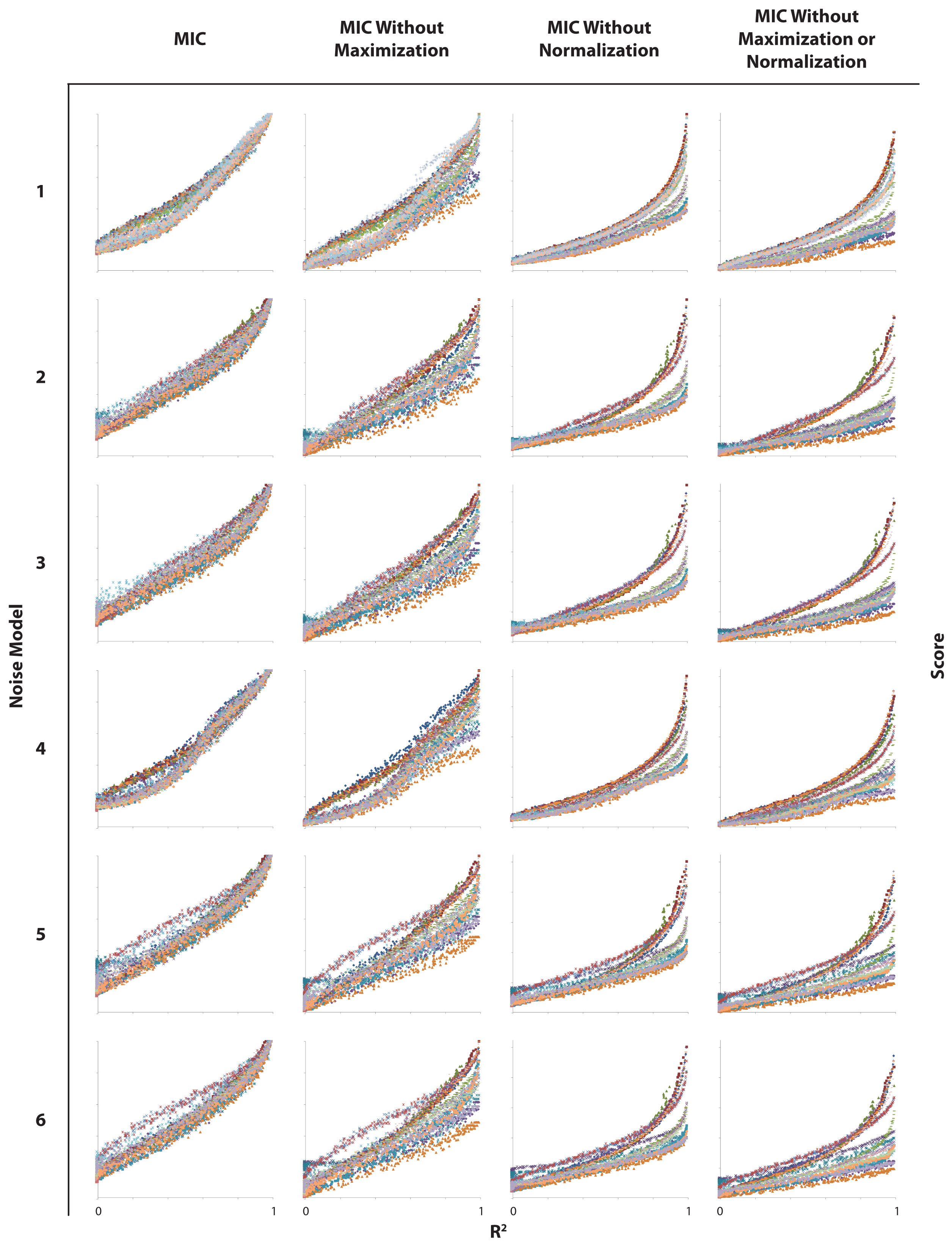}
        \caption[The role of maximization and normalization in the definition of MIC]{The behavior of MIC and three variants of MIC defined in Section~\ref{sec:intuition} on the noisy functional relationships discussed in Section~\ref{sec:Preliminaries}.  Each plot contains the score of the given statistic versus the coefficient of determination, $R^2$, of the noisy data relative to the noiseless function in question.  For noise model 1, the plot legend is the same as that presented in Figure~\ref{EquitabilityIntro}; for noise models 2-6, the legend of relationship types and sample sizes is presented in Appendix~\ref{appendixA} (Figure~\ref{appendix:SingleTrialPlotLegends}).  From left to right in the figure, the three variants of MIC used were $\MIC_1$ (range: $[0,1]$), $\MIC_2$ (range plotted: $[0,2.6]$), and $\MIC_3$ (range plotted: $[0,2.6]$).  All three variants produce non-equitable behavior, demonstrating that it is the combination of maximization and normalization that leads to the equitability of MIC.}
\label{EffectOfMaxAndNormInMIC}
\end{figure}

Figure~\ref{EffectOfMaxAndNormInMIC} shows that MIC provides substantially greater equitability than any of the three variants defined above. The figure contains MIC scores plotted against $R^2$ scores for each relationship in the suite of test functions used in \citet{MINE} and listed in Table 1. Each data point corresponds to an independent realization of noisy function data with a given noise level. Visually then, equitability corresponds to how tightly coupled the points are. More directly, the property we seek is that for many fixed scores of the statistic being tested, the range of $R^2$ scores of data that received each score is small.

A related but stronger property that we might want is that the scores given by the statistic being tested should track the noise level; that is, we want the statistic being evaluated to roughly equal $R^2$ as the noise changes.  While not implied by equitability, this property allows an equitable statistic to be interpreted even more intuitively. MIC achieves this stronger property, as well as equitability, much more effectively than the variants introduced above.

The results in Figure~\ref{EffectOfMaxAndNormInMIC} demonstrate that both the maximization and the normalization in the definition of MIC are necessary for its equitability. Without the normalization, relationships that are better captured by grids with more cells are favored over those that are better captured by simple grids. For instance, the sinusoids, which are best captured by grids with 2 rows and many columns, will never have scores above $\log_2(2) = 1$, while the more monotonic relationships can more easily be captured by grids with both many rows and many columns, and so they achieve scores above 1. The maximization step also proves necessary for equitability: without it, relationships that are not naturally equipartitioned are unduly penalized. For example, while ordinary sinusoids are well captured by equipartitions, the varying-wavelength sinusoids are not, causing them to receive lower scores.

\section{The Approximation Algorithm for MIC}

The approximation algorithm given in \citet{MINE} for computing MIC has several parameters, two of which we focus on here. The first is the exponent alpha in the function $B(n) = n^\alpha$. The second is $c$, which gives a speed versus optimality tradeoff by specifying how fine-grained the search for the optimal grid is: roughly, if a grid with $y$ columns is sought, the algorithm first creates an equipartition with $cy$ columns and then searches for an optimal sub-partition containing only $y$ columns.   The above simulations were generated using the default values of $\alpha$ and $c$: $\alpha=0.6$ and $c=15$.

As sample sizes grow, we have found that changing the values of $\alpha$ and $c$ from these defaults can significantly speed up the algorithm with little effect on the equitability of the resulting statistic. As an example, Table~\ref{table:Runtimes} compares the runtime of the algorithm using the recommended parameters from \citet{MINE} to a different setting of $\alpha=0.55$, $c=5$. This latter setting is much faster and does not appear to significantly affect performance. To emphasize this point, we use the new parameters, $\alpha = 0.55$ and $c=5$, for the remainder of this paper.

We note that for samples sizes beyond the regimes addressed here the parameter $\alpha$ can be further reduced. This is because $B(n) = n^\alpha$ governs the maximal ``complexity" of the relationships found by MINE (i.e. the number of cells required to effectively describe the relationships using a grid). In many cases there is an upper limit in practice on the complexity that is sought, and so $B(n)$ need not grow much beyond a certain point.

\begin{table}
\begin{center}
\begin{tabular}{l c c c c c c c c}
\textbf{\scriptsize Sample} &\textbf{\scriptsize Line} & \textbf{\scriptsize Exp.} & \textbf{\scriptsize Sigmoid} & \textbf{\scriptsize Parabola} & \textbf{\scriptsize Cubic} & \textbf{\scriptsize Sine} & \textbf{\scriptsize Varying Freq.} & \textbf{\scriptsize Random}\\
\textbf{\scriptsize Size} & & \textbf{\scriptsize $[10^x]$} &  &  &  & \textbf{\scriptsize (High Freq.)} & \textbf{\scriptsize Cosine} & \\
\hline
n=200 & 25 & 20 & 21 & 25 & 26 & 28 & 27 & 32 \\
n=400 & 92 & 88 & 89 & 111 & 112 & 113 & 117 & 135 \\
n=600 & 190 & 194 & 191 & 245 & 246 & 253 & 249 & 301 \\
n=800 & 361 & 362 & 353 & 444 & 453 & 475 & 481 & 553 \\
n=1000 & 551 & 548 & 565 & 693 & 714 & 728 & 739 & 857 \\
n=2000 & 2111 & 2101 & 2135 & 2541 & 2580 & 2663 & 2680 & 3024 \\
n=4000 & 8178 & 7892 & 8063 & 9626 & 9863 & 10076 & 10170 & 11279 \\
n=6000 & 17081 & 16641 & 17022 & 20187 & 20620 & 21087 & 21193 & 23517 \\
n=8000 & 28982 & 28037 & 28882 & 33989 & 34682 & 35561 & 35752 & 39362 \\
n=10000 & 43465 & 42315 & 43264 & 51041 & 52272 & 53780 & 53885 & 59123 \\
\hline \hline
*n=200 & 6 & 2 & 2 & 2 & 2 & 2 & 2 & 2 \\

*n=400 & 7 & 5 & 6 & 6 & 6 & 7 & 7 & 7 \\

*n=600 & 14 & 11 & 12 & 13 & 14 & 14 & 14 & 15 \\

*n=800 & 21 & 19 & 19 & 22 & 22 & 23 & 23 & 25 \\

*n=1000 & 31 & 28 & 28 & 33 & 33 & 35 & 35 & 37 \\

*n=2000 & 93 & 89 & 91 & 105 & 107 & 112 & 110 & 118 \\

*n=4000 & 284 & 277 & 287 & 322 & 330 & 347 & 341 & 363 \\

*n=6000 & 566 & 545 & 555 & 637 & 652 & 679 & 673 & 716 \\

*n=8000 & 944 & 911 & 935 & 1082 & 1082 & 1147 & 1129 & 1201 \\

*n=10000 & 1325 & 1297 & 1335 & 1512 & 1541 & 1623 & 1604 & 1706 \\
\hline
\end{tabular}
\caption[Run-times, in milliseconds, of the algorithm for generating the characteristic matrix and calculating MIC, using both default and modified parameters, on a range of functional relationships, noise levels, and sample sizes.]{Run-times, in milliseconds, of the algorithm for generating the characteristic matrix and calculating MIC, using both default and modified parameters, on a range of functional relationships, noise levels, and sample sizes.  The top half of the table corresponds to the default parameters $\alpha=0.6$ and $c=15$, while the bottom half (indicated by *) was run with the modified parameters $\alpha=0.55$ and $c=5$.  For each sample size and relationship type, the average run-time in milliseconds over a range of 10 noise levels interpolated between $R^2=1.0$ and $R^2=0.0$ is presented. As sample sizes grow, changing the parameters $\alpha$ and $c$ significantly lowers the run-time of the algorithm.}
\label{table:Runtimes}
\end{center}
\end{table}

\begin{figure}
	\centering
	\includegraphics[trim = 1in 6.85in 0.85in 1.35in, clip, width = \textwidth]{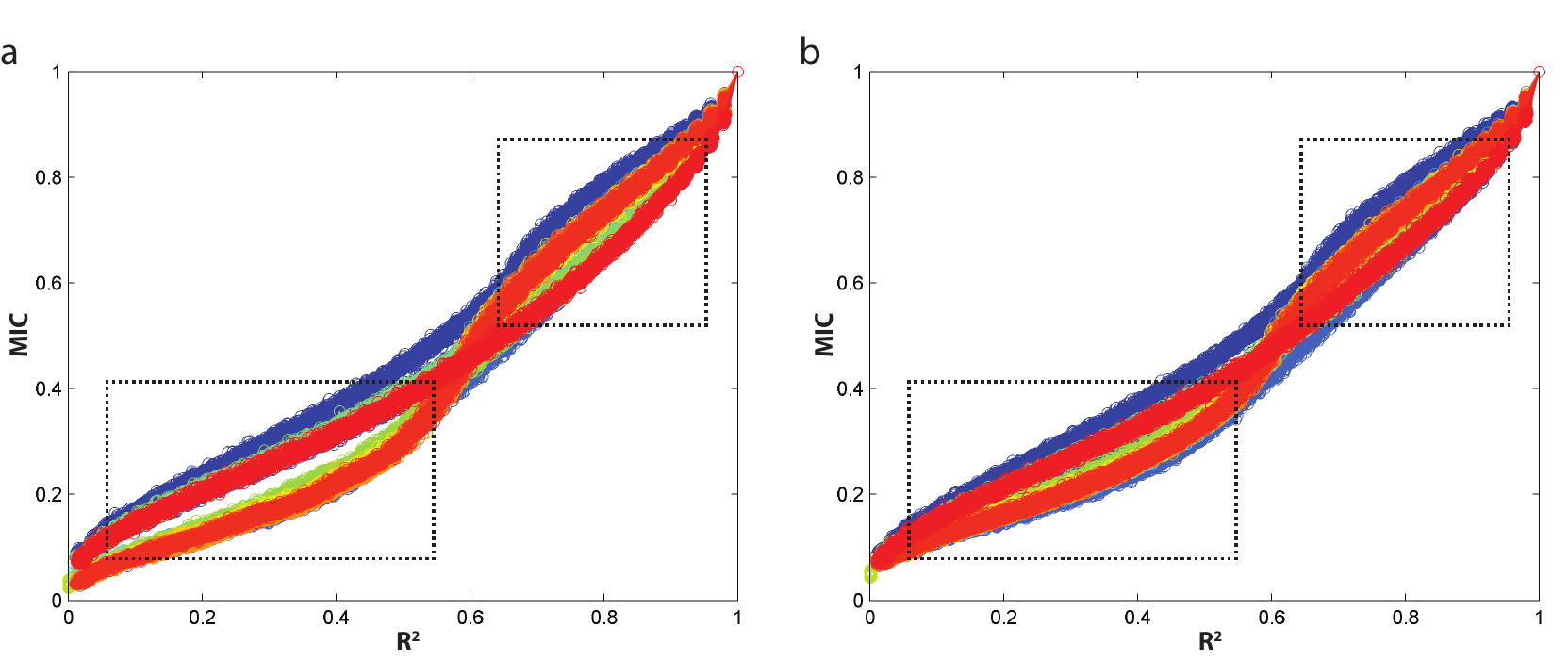}
	\caption[Improving the algorithm for approximating MIC improves the equitability of reported MIC values]{Improving the algorithm for approximating MIC improves the equitability of reported MIC values. \textbf{(a)} 100 overlaid iterations of the equitability analysis performed in Figure~\ref{EquitabilityIntro} using noise model 4 and $n=5000$, and using the standard algorithm for approximating MIC. The colors corresponding to each type of functional relationship are listed in Appendix~\ref{appendixA} (Figure~\ref{appendix:ManyTrialPlotLegends}).  \textbf{(b)} The same equitability analysis carried out using a modified, more computationally intensive algorithm, which comes closer to computing the true value of MIC.  When this algorithm is used, the equitability of MIC improves, as demonstrated by the diminishing of the gap in MIC scores across different types of functional relationships outlined by the boxes.}
\label{betterApproxAlgorithm}
\end{figure}

We also examined whether use of the approximation algorithm from \citet{MINE} affects equitability in comparison to a less efficient algorithm that more exhaustively searches for optimal grids. To do this, we modified the original algorithm presented in \citet{MINE} such that for all grids with 2 or 3 rows, it no longer simply equipartitions the $y$-axis but rather exhaustively searches an equipartition of the $y$-axis into 20 rows in order to find the best subpartition into 2 or 3 rows respectively. As Figure~\ref{betterApproxAlgorithm} shows, this more exhaustive algorithm has better equitability than the original approximation algorithm presented in \citet{MINE}, suggesting that some of the deviations from equitability of currently reported MIC values are not intrinsic to MIC but rather introduced by the approximation algorithm used to compute it. We expect that approximation algorithms with better time-accuracy tradeoffs may be found with further study.

\section{The Tradeoff between Equitability and Power}

While MIC has the advantage of equitability, which allows it to pick out the strongest relationships in a data set, it has lower power than other methods for detecting as many weak relationships as possible \citep{MINE, simon2012comment, gorfine2012comment}. To explore this tradeoff between equitability and power, we contrast MIC with distance correlation, an elegant measure of dependence introduced by Szekely and Rizzo \citep{szekely2009brownian}.

\begin{figure}
	\centering
	\subfloat {\includegraphics[trim = 0in 5.5in 0in 0in, clip, width = \textwidth]{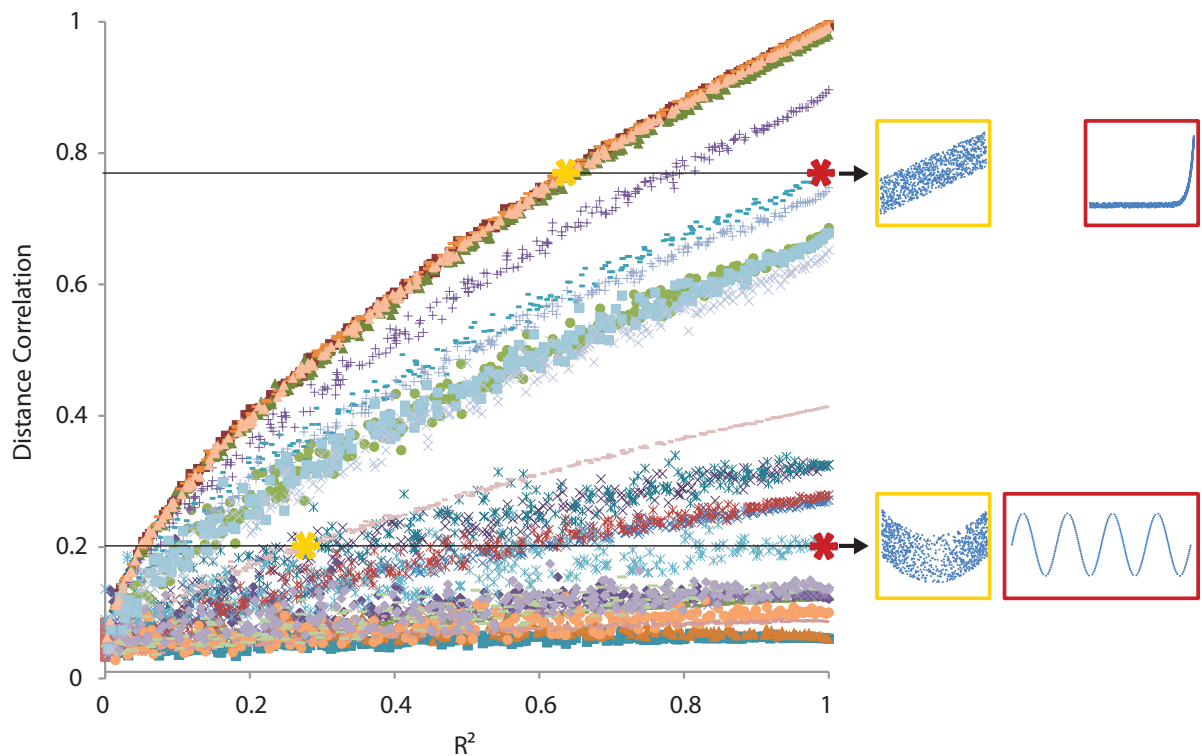}}
	\caption[Lack of equitability of distance correlation on noisy functional relationships]{The lack of equitability of distance correlation on a test suite of noisy functional relationships. The plot contains the distance correlation of 27 different functional relationships with various sample sizes and increasing amounts of noise plotted against the coefficient of determination ($R^2$) of each relationship relative to its generating function. Noise was generated using noise model 1 (points spaced evenly along the curve described by the function, uniform vertical noise). Thumbnails shown to the right of the plot show relationships that received identical scores. This plot demonstrates that distance correlation is highly non-equitable on functional relationships, giving similar scores to relationships with widely varying noise levels.  The legend for the functions used is the same as that provided in Figure~\ref{EquitabilityIntro}.\\

\textit{Reproduced from Figure S3 of the Supplemental Online Material of \citet{MINE}.}}
\label{MINE_FigS3}
\end{figure}

\begin{figure}
        \centering
	  \includegraphics[trim = 0in .05in 0.6in .05in, clip, width = 0.78\textwidth]{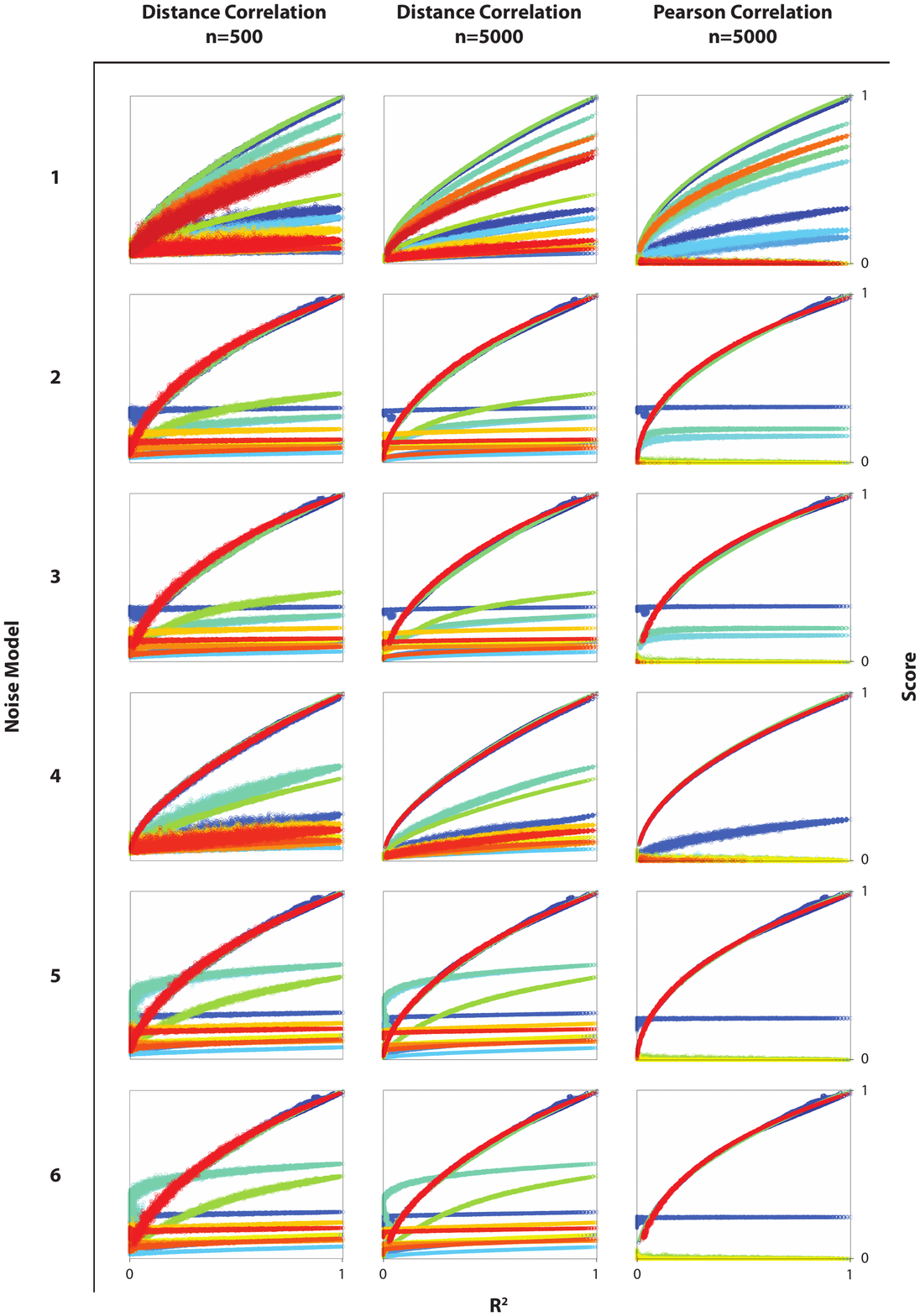}
        \caption[Equitability and variance of distance correlation on noisy functional relationships with $n=500$ and $n=5000$]{Equitability and variance of distance correlation on noisy functional relationships with $n=500$ and $n=5000$.  Plots contain the distance correlation of the suite of functional relationships described in Table~\ref{table:FctTestSuite} with increasing amounts of noise plotted against the coefficient of determiniation, $R^2$. To reflect the effect of the variance of distance correlation on its equitability, plots contain 100 independent realizations of each noisy relationship (each relationship type is colored differently; legend in Figure~\ref{appendix:ManyTrialPlotLegends} of Appendix~\ref{appendixA}). Noise was generated using the six different noise models described in Section~\ref{sec:Preliminaries}. Distance correlation is highly non-equitable across all the noise models tested, and its equitability profile is similar to that of the Pearson correlation (analysis provided only for $n=5000$).}
\label{DcorEquitabilityAnalysis}
\end{figure}

Distance correlation belongs to a large class of methods designed for testing for the \em presence \em of statistical dependence. This is a fundamentally different problem than the one posed in \citet{MINE}: quantifying the \em strength \em of a dependence in order to identify a small set of strongest associations in a data set. Thus, on the one hand distance correlation indeed has better power than MIC for many relationship types \citep{simon2012comment, gorfine2012comment}.\footnote{It is important to note that this does not affect the false positive rate of MIC if p-values are calculated appropriately.  Given the empirical distribution of MIC scores, it is possible to determine the proper cutoff for testing the independence hypothesis for a given, desired false positive rate. (\citet{MINE} provide these cutoffs for a range of sample sizes and desired false positive rates.) If this is done, decreased power manifests itself in a decreased ability to detect weak relationships rather than in an increased false positive rate.} On the other hand, however, distance correlation is highly non-equitable across all the noise models tested, and in fact its equitability profile is similar to that of the classical Pearson product-moment correlation. This is shown in Figure S3 of \citet{MINE} (reproduced here as figure~\ref{MINE_FigS3}) as well as in Figure~\ref{DcorEquitabilityAnalysis}.

While a method with better power is always preferable if all other things are equal, the desiderata of a statistic depend on the problem it is being used to solve, and distance correlation's lack of equitability makes it ill-suited for the data exploration setting posed in \citet{MINE}.  MIC is a more appropriate measure of dependence in a situation in which there are likely to be an overwhelming number of significant relationships in a data set, and there is a need to automatically find the strongest ones.

\section{Comparison to Mutual Information}

Given that mutual information appears in the definition of MIC itself, it is natural to ask whether direct estimation of mutual information yields an equitable statistic. The answer to this question appears generally to be `no'.  Here we provide evidence for this claim beyond that given in \citet{MINE}, considering a range of data set sizes and noise models as well as different parameters to the methods used for estimating MIC and mutual information.

When using direct mutual information estimation for data exploration, it is useful to normalize the resulting scores in order to obtain a measure between 0 and 1. Here we have used the squared Linfoot correlation, defined as $1-2^{-2I}$, where $I$ is the mutual information of a relationship \citep{linfoot1957informational, speed2011correlation, kinney2012}.  With this normalization, a score of 0 represents a mutual information of 0 (i.e. statistical independence) while a score of 1 represents a mutual information of infinity.

An additional consideration is how to estimate mutual information given a finite set of samples.  In \citet{MINE}, the well-known Kraskov et al. estimator was used with the smoothing parameter $k$ (which governs the number of nearest neighbors used for each point in the computation) set to the standard default value of $k=6$.  A possible alternative is to set this parameter to the minimal possible value of $k=1$, in order to minimize bias \citep{kinney2012}.

\begin{figure}
        \centering
        \includegraphics[trim = 0in .075in 0.6in .05in, clip, width = 0.77\textwidth]{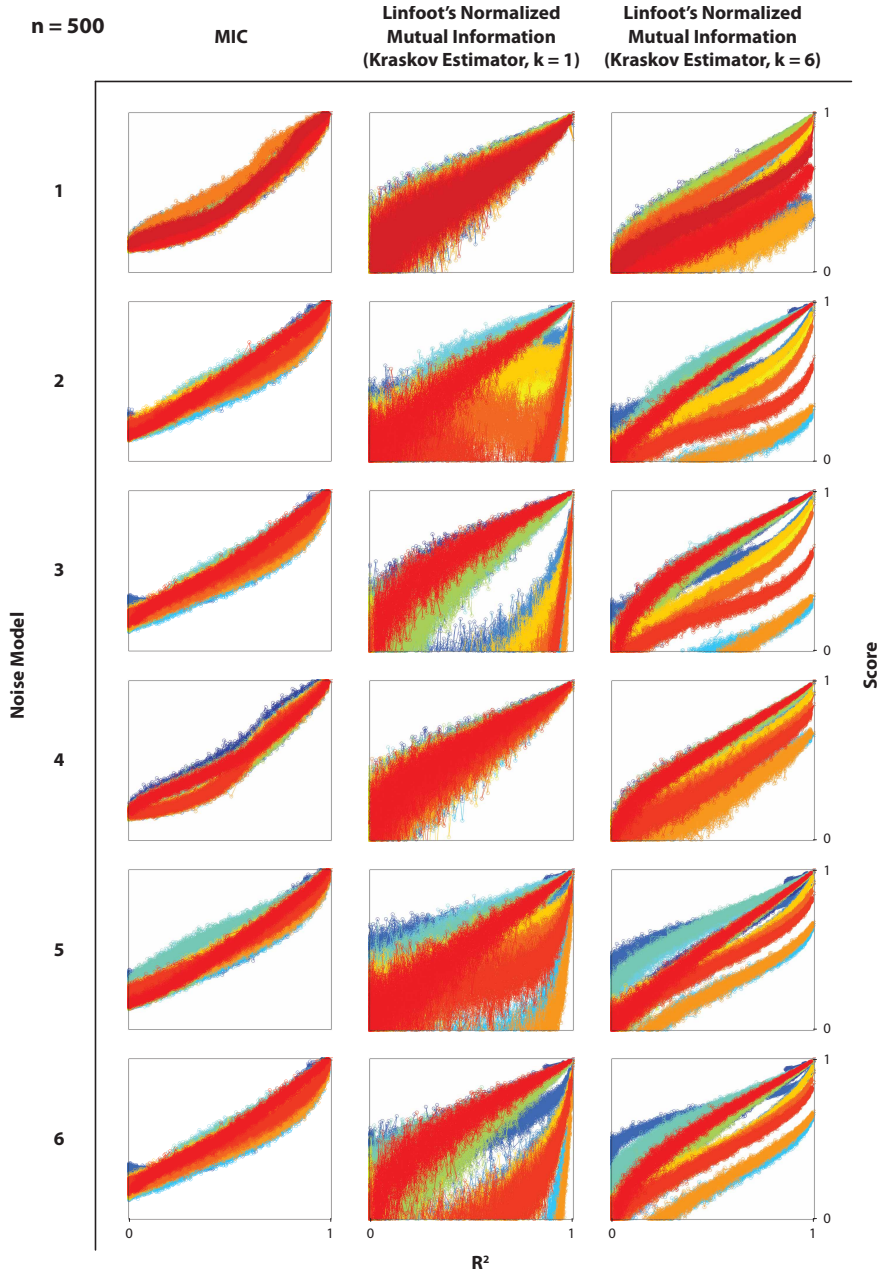}
        \caption[Equitability and variance of MIC and mutual information (squared Linfoot correlation) on noisy functional relationships with $n=500$]{Equitability and variance of MIC and mutual information (squared Linfoot correlation) on noisy functional relationships with $n=500$. Each plot contains mutual information (estimated using Kraskov et al. estimator with $k=1$ and $k=6$) and MIC scores of the suite of functional relationships ($n=500$) described in Table~\ref{table:FctTestSuite} with increasing amounts of noise plotted against the coefficient of determination, $R^2$. To reflect the effect of the variance of each method on its equitability, plots contain 100 independent realizations of each noisy relationship (each relationship type is colored differently; legend in Figure~\ref{appendix:ManyTrialPlotLegends} of Appendix~\ref{appendixA}). Noise was generated using the six different noise models described in Section~\ref{sec:Preliminaries}. At $n=500$, MIC outperforms mutual information in terms of equitability under all noise models considered, and regardless of the choice of smoothing parameter used in the Kraskov et al. estimator.}
\label{MIVarianceAnalysis500}
\end{figure}

\begin{figure}
        \centering
        \includegraphics[trim = 0in .075in 0.6in .05in, clip, width = 0.77\textwidth]{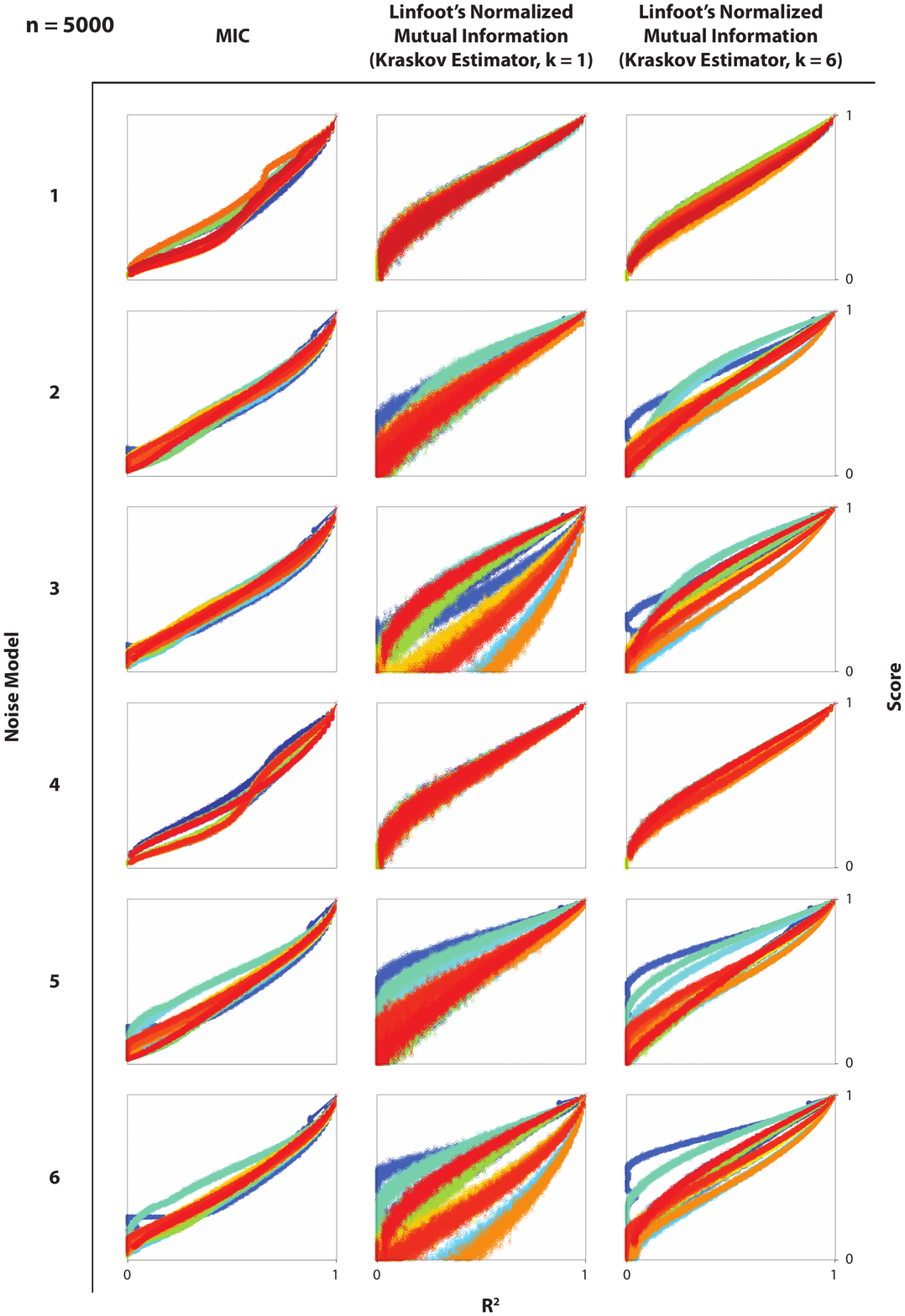}
        \caption[Equitability and variance of MIC and mutual information (squared Linfoot correlation) on noisy functional relationships with $n=5000$]{Equitability and variance of MIC and mutual information (squared Linfoot correlation) on noisy functional relationships with $n=5000$. Each plot contains mutual information (estimated using Kraskov et al. estimator with $k=1$ and $k=6$) and MIC scores of the suite of functional relationships ($n=5000$) described in Table~\ref{table:FctTestSuite} with increasing amounts of noise plotted against the coefficient of determination, $R^2$. To reflect the effect of the variance of each method on its equitability, plots contain 100 independent realizations of each noisy relationship (each relationship type is colored differently; legend in Figure~\ref{appendix:ManyTrialPlotLegends} of Appendix~\ref{appendixA}). Noise was generated using the six different noise models described in Section~\ref{sec:Preliminaries}. At $n=5000$, MIC outperforms mutual information in terms of equitability under all noise models containing horizontal noise, and regardless of the choice of smoothing parameter used in the Kraskov et al. estimator.}
\label{MIVarianceAnalysis5000}
\end{figure}

Using the test suite of functions provided in \citet{MINE}, we compare MIC to mutual information (squared Linfoot correlation) estimated using the Kraskov et al. estimator with $k=1$ (minimal smoothing) and $k=6$ (the default setting), under all six noise models discussed above, and at sample sizes of $n=500$ and $n=5000$ (Figures~\ref{MIVarianceAnalysis500} and~\ref{MIVarianceAnalysis5000}). While $n=5000$ may be a larger sample size than would be realistically obtained in many experimental settings, we examine it because the Kraskov et al. mutual information estimator performs better under larger sample sizes. In these analyses, as in Figure~\ref{DcorEquitabilityAnalysis}, we repeat each experiment 100 times so that the plots capture the effect of the variance of each estimator on its equitability.

As Figures~\ref{MIVarianceAnalysis500} and~\ref{MIVarianceAnalysis5000} show, the Kraskov mutual information estimator with $k=1$ has a very high variance. That is, the scores given by the estimator to independent realizations of the same relationship (i.e. independent, identically sized sets of samples from the same distribution) themselves vary widely under this setting. This result is consistent with remarks made in \citet{Kraskov} discouraging the use of $k=1$ for this reason. The high variance of the estimator here naturally results in poor equitability: before an equitable statistic gives similar scores to similarly noisy relationships of different types, it needs to be able to give similar scores to similarly noisy relationships of the same type. 

With $k=6$, mutual information is significantly less equitable than MIC across all the noise models tested in the $n=500$ sample size regime. And at $n=5000$, MIC likewise outperforms mutual information on most noise models, with the case of vertical noise alone being the only setting where the schemes appear comparable. For instance, under the noise models that include both horizontal and vertical noise, the difference in mutual information scores of relationships in the test suite with identical $R^2$ values reaches 0.65, even at a sample size of $n=5000$. And under the two noise models that include horizontal noise only, the difference in scores of relationships with identical $R^2$ values reaches 0.88. These behaviors persist at sample sizes of $n=10000$ and $n=20000$ as well, suggesting that they are due not just to potential bias of the Kraskov et al. estimator, but also to the properties of mutual information itself.

\section{Conclusion}

Our analysis shows that, under most noise models and sample sizes, the normalization and maximization steps involved in computing MIC are necessary for its equitability, and that these elements make MIC more equitable than mutual information estimation. This was shown both by measuring the equitability of variations on MIC with each of these features removed, as well as by comparing MIC to the Kraskov et al. mutual information estimator under six different noise models and at sample sizes of $n=500$ and $n=5000$. 

Our work here suggests that at larger sample sizes, the default parameters given in \citet{MINE} can be modified to gain a significant decrease in runtime without significant loss of equitability. Our analyses also show that at least some meaningful part of the deviation from equitability of currently reported MIC values is due to errors introduced by the current approximation algorithm rather than the intrinsic behavior of MIC. Both of these issues appear worthy of further study for both theoretical understanding and practical improvements of MIC's performance.  In particular, we hope that better, faster approximation algorithms will arise with further research \citep{albanese2012cmine}.

Equitability is one of arguably many examples where one might want to measure some property of a relationship that is simple to compute given knowledge of the relationship type, but non-trivial to measure without that knowledge. We can call such statistics \em class-independent\em, because they do not require fore-knowledge of the class of the relationship (e.g. linear, exponential, non-functional, etc.) under consideration. For instance, in this framework an equitable statistic for noisy functional relationships would be a class-independent measure of $R^2$. An interesting direction of future work would be to define other desirable class-independent statistics and find efficient ways to compute them.

\section{Acknowledgments}
We would like to thank H. Finucane for valuable discussions and helpful suggestions throughout. This work was supported in part by NSF grants NSF IIS-0964473 and NSF CCF-0915922 (M.M.), the Paul and Daisy Soros Foundation (D.N.R), and the Packard Foundation (P.C.S.).

\newpage

\appendix
\section{Appendix A}
\label{appendixA}

\begin{figure}
	\centering
	\includegraphics[trim = 0in 5.5in 3in 0in, clip, width = \textwidth]{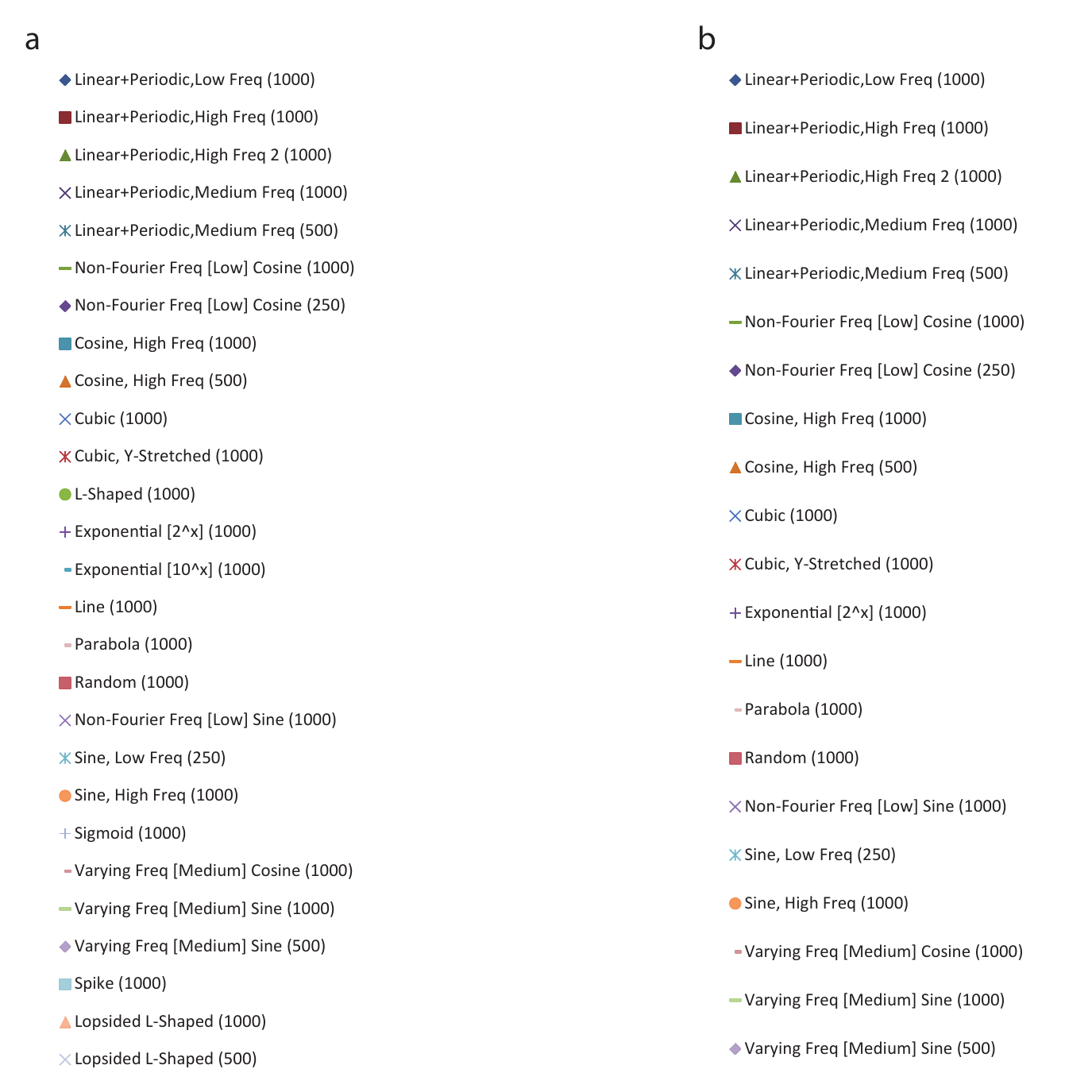}
	\caption[Legend of the suite of functional relationships and sample sizes used in analyses in Figures~\ref{EquitabilityIntro},~\ref{EffectOfMaxAndNormInMIC}, and~\ref{MINE_FigS3}]{Legend of the suite of functional relationships and sample sizes used in analyses in Figures~\ref{EquitabilityIntro},~\ref{EffectOfMaxAndNormInMIC}, and~\ref{MINE_FigS3}. (a) The legend for analyses performed using noise model 1. (b) The legend for analyses performed using noise models 2,3,4,5, and 6.  All function names refer to those used in Table~\ref{table:FctTestSuite} and numbers in parentheses are sample sizes.}
\label{appendix:SingleTrialPlotLegends}
\end{figure}

\begin{figure}
	\centering
	\includegraphics[trim = 0in 6.25in 2in 0in, clip, width = \textwidth]{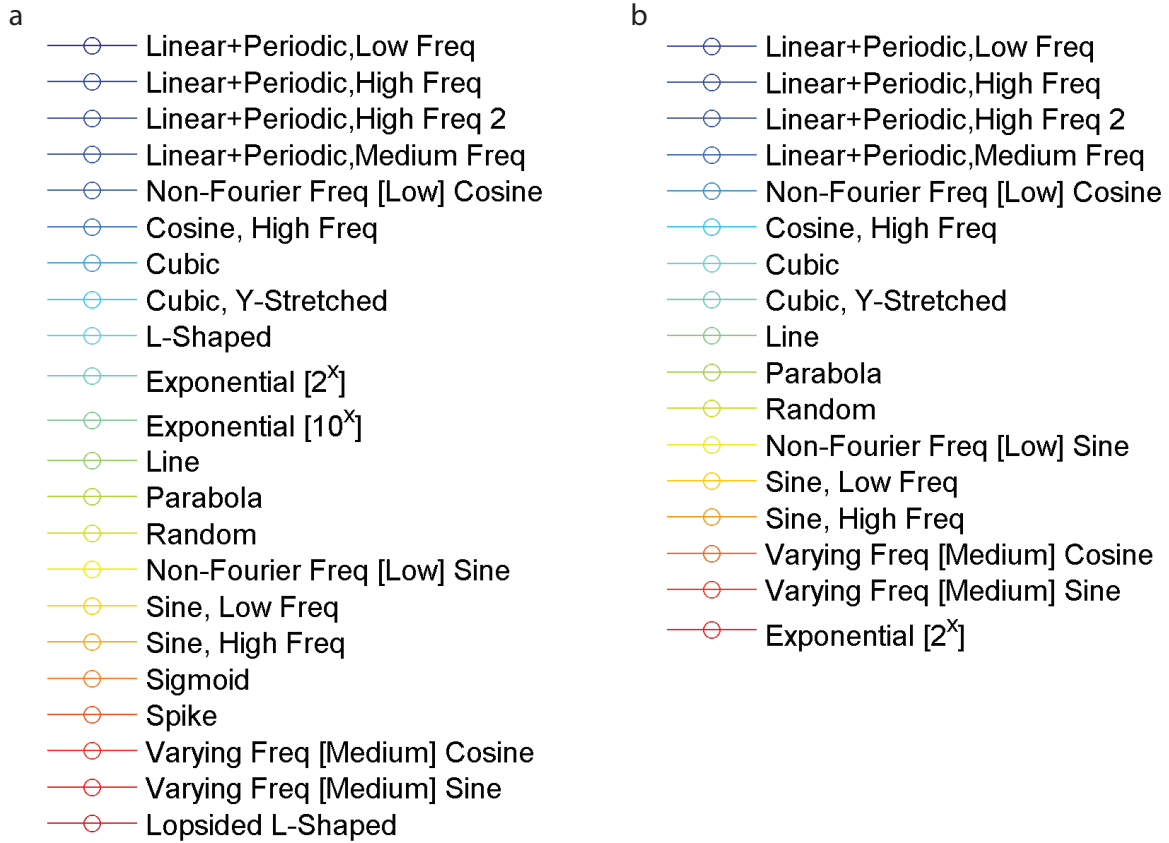}
	\caption[Legend of the suite of functional relationships used in analyses in Figures~\ref{betterApproxAlgorithm},~\ref{DcorEquitabilityAnalysis},~\ref{MIVarianceAnalysis500}, and~\ref{MIVarianceAnalysis5000}]{Legend of the suite of functional relationships used in analyses in Figures~\ref{betterApproxAlgorithm},~\ref{DcorEquitabilityAnalysis},~\ref{MIVarianceAnalysis500}, and~\ref{MIVarianceAnalysis5000}. (a) The legend for analyses performed using noise model 1. (b) The legend for analyses performed using noise models 2, 3, 4, 5, and 6.  All function names refer to those used in Table~\ref{table:FctTestSuite}.}
\label{appendix:ManyTrialPlotLegends}
\end{figure}

This appendix contains legends for the plots presented in the figures above.  Figure~\ref{appendix:SingleTrialPlotLegends} contains a legend of the suite of functional relationships and sample sizes used in Figures~\ref{EquitabilityIntro},~\ref{EffectOfMaxAndNormInMIC}, and~\ref{MINE_FigS3}, and Figure~\ref{appendix:ManyTrialPlotLegends} contains a legend of the suite of functional relationships used in analyses in Figures~\ref{betterApproxAlgorithm},~\ref{DcorEquitabilityAnalysis},~\ref{MIVarianceAnalysis500}, and~\ref{MIVarianceAnalysis5000}.  In both figures, there is a distinction between the set of functional relationships that are used when analyses were performed using noise model 1 and the set that are used for analyses using noise models 2, 3, 4, 5, and 6.  When considering noise models 2, 3, 4, 5, and 6, functions with very steep portions are omitted and the ``Exponential [$2^x$]'' function has $x \in [0,2]$ rather than $x \in [0,10]$. This is because adding horizontal noise to a steep function distorts its $R^2$, and because sampling uniformly along the $x$-axis for steep functions made them appear discontinuous.

\newpage
\bibliographystyle{plainnat}
\bibliography{References}

\end{document}